\documentclass[journal]{IEEEtran}

\usepackage{graphicx, epstopdf}
\usepackage{setspace}
\usepackage{acronym}
\usepackage{amssymb}
\usepackage[cmex10]{amsmath}
\usepackage{commath}
\usepackage{mathrsfs}
\usepackage{mathtools}
\usepackage{ifthen}
\usepackage{textcomp}
\usepackage{float}
\usepackage[tight,footnotesize]{subfigure}
\usepackage{cite}
\usepackage{bm}
\usepackage{url}
\usepackage{color,soul}
\usepackage{siunitx}
\input{insbox}

\IEEEoverridecommandlockouts

\begin{document}

\title{Automatic Quantification of Settlement Damage using Deep Learning of Satellite Images}

\author{Lili Lu\textsuperscript{1}, Weisi Guo\textsuperscript{1,2*}

\thanks{\textsuperscript{1}SATM, Cranfield University. \textsuperscript{2}Alan Turing Institute. *Corresponding Author: weisi.guo@cranfield.ac.uk. }}

\maketitle

\begin{abstract}
Humanitarian disasters and political violence cause significant damage to our living space. The reparation cost to homes, infrastructure, and the ecosystem is often difficult to quantify in real-time. Real-time quantification is critical to both informing relief operations \cite{4215094, 7892890, 8114334, 8585069}, but also planning ahead for rebuilding. Here, we use satellite images before and after major crisis around the world to train a robust baseline Residual Network (ResNet) and a disaster quantification Pyramid Scene Parsing Network (PSPNet). ResNet offers robustness to poor image quality and can identify areas of destruction with high accuracy (92\%), whereas PSPNet offers contextualised quantification of built environment damage with good accuracy (84\%). As there are multiple damage dimensions to consider (e.g. economic loss and fatalities), we fit a multi-linear regression model to quantify the overall damage. To validate our combined system of deep learning and regression modeling, we successfully match our prediction to the ongoing recovery in the 2020 Beirut port explosion. These innovations provide a better quantification of overall disaster magnitude and inform intelligent humanitarian systems of unfolding disasters.
\end{abstract}

\section{Introduction}
Rapid advances in deep learning (DL) has enabled significant remote sensing applications in automated systems for disaster management. According to the World Health Organization (WHO), 90,000 people die 160 million people are severely affected by a variety of natural disasters. Furthermore, 45 million people are displaced from their homes due to a variety of natural disasters and political violence. In the whole year of 2019, there were 409 severe natural disasters occurred worldwide. Therefore, there is clearly a challenge to quantify the magnitude of said disasters and conflicts, in order to inform a range of government and NGO systems to provide appropriate assistance and recovery plans.

\subsection{Automated Disaster Quantification and Management}
In post disaster scenarios, site exploration and disaster evaluation is challenged by dangerous and difficult environments for first responders. Scalable quantification, updated quantification is extremely challenging over vast areas. For example Hurricane Katrina (2005) caused damage over 230,000 km$^{2}$ grossing to over \$160bn of damage. 

Satellite images have been previously used for understanding site access and automating responses, but manual feature processing is often slow and lacks scalability to large areas. Recent advances in deep learning can identify of hazard zones, provide risk assessment and insurance compensation. The data used includes a combination of aerial drone photography, satellite imaging, and social media data. In recent years, automated disaster detection system \cite{Amit16} have used CNNs to extract data from the immediate disaster area and prioritise resource distribution and recognise viable infrastructure for aid delivery \cite{RSOS}. The results show that the accuracy of disaster detection is 80-90\%. Similar work using aerial photos from UAVs have been used with a VGGNet deep learning model achieving 91\% accuracy \cite{Kamilaris18}. Recognising the complex discontinuities in disaster images, improvements were made by adding a residual connection and extended convolution to the previous CNN frameworks \cite{Duarte18}. When combined with the feature maps generated by aerial images and satellite image samples, the work improved the overall classification of the satellite images for building damage by nearly 4\%. More recently, researchers have developed a new deep learning approach to analyse flood disaster images and quickly detect areas that have been flooded or destroyed to assess the extent and severity of damage \cite{Sublime19}. However, the open challenge in this area is \textit{scene parsing}, whereby recognising and segmenting livable objects (e.g. houses, infrastructure) from natural environment (e.g. trees and open spaces) is critical to quantifying the built livable space impact. Another open challenge is that quantification thus for in the above work has focused on singular dimensions, however we must recognise that economic damage and human fatality are both important indicators. Therefore, we would need to develop a model that combines them.

\begin{figure*}[t]
	\centering
	\includegraphics[width=1.0\linewidth]{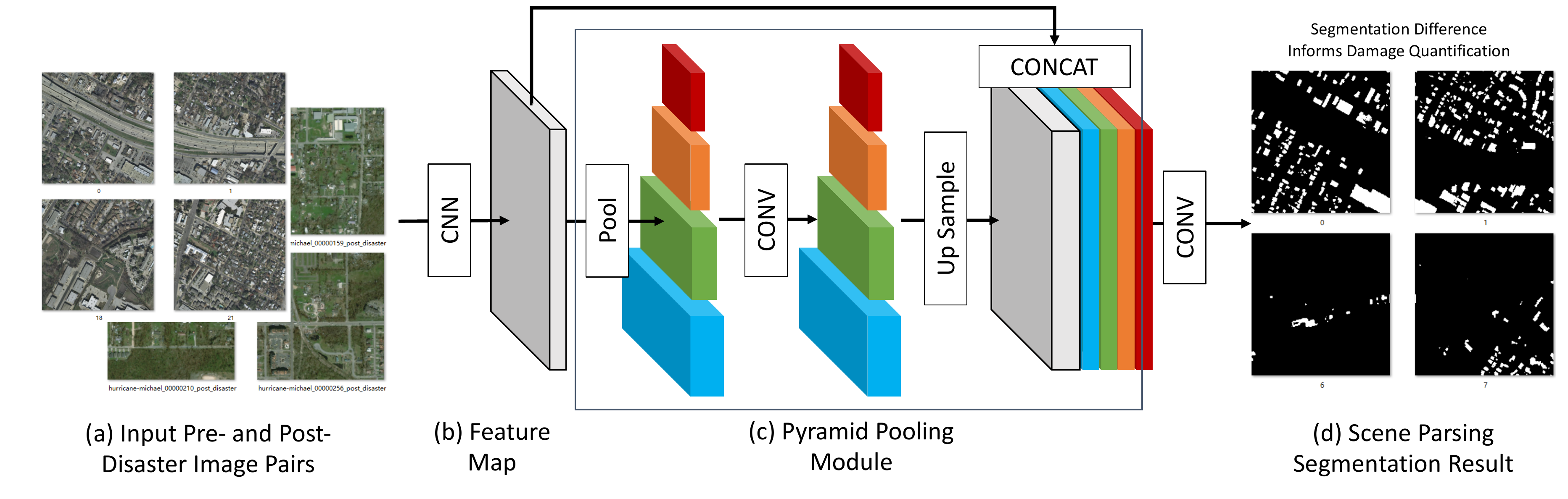}
	\caption{Disaster scene parsing using PSPNet: (a) pre- and post-disaster images containing both built and natural environment, (b-c) feature map feeds a pyramid pooling module to achieve (d) contextual object segmentation that informs damage quantification.}
	\label{1}
\end{figure*}

\subsection{Contribution}
This paper aims to find a way to describe the severity of the disaster and try to quantify the disaster automatically. Our contributions and novelties are as follows:

(1) We use over 100 case studies from open sources, which includes before and after photos, disaster information (e.g. date, cost, fatality, etc.) and created a novel database which previously did not exist before for academia. 

(2) We compare two state-of-the-art approaches in deep learning, the readily available Residual Network (ResNet) and the more suitable custom configured Pyramid Scene Parsing Network (PSPNet). The former ResNet has a lower requirement for image quality, whereas the latter PSPNet is capable of scene parsing for livable space separation from natural environment. We compare the trade-off performance of image quality requirement vs. damage quantification accuracy.

(3) We develop a multi-linear regression model to map the neural network outputs from (2) to the actual economic and human damage values. We keep (2) and (3) separate in order to enable the neural network models in (2) to be useful to a much wider set of humanitarian contexts beyond damage quantification. 

(4) Finally we train our mode on historical cases across a spectrum of global  disasters and then test our model on a more recent event.

As far as we are aware of, no such combined deep learning and disaster quantification system exists. As far as we are aware of, no such combined deep learning and disaster quantification system exists. The system fits into a wider need to automate humanitarian disaster response, allocate and prioritise global emergency resources to either preemptively address or respond to disasters \cite{Guo19}.

\section{Data and Methodology}

\subsection{Database}
We source all our images and accompanying disaster data from open source websites. The images should have high quality (256×256 dpi) and 124 disasters are covered ranging from forest fire, earthquake, mudslide, tsunami, volcanic eruption, hurricane, typhoon, tornado and major industrial explosions, spanning several recent years around the world. Our major sources of data are: (1) EM-DAT international disaster database established by WHO and the Belgian Government \cite{EM-DAT}, (2) CRED Centre for Research on the Epidemiology of Disasters, (3) Inria Aerial Image Dataset \cite{Maggiori17}, and (4) xBD dataset \cite{Gupta19}. We summarize our data in Table 1.

\begin{table}
\small
\centering
\caption{Data Parameters}
\label{Table1}
\begin{tabular}{cc}
\hline
Data or Parameter        & Value \\ \hline
Case Studies             & 124 \\
Disaster Categories      & 9 \\
Time Span                & 2004 to 2019 \\
Location                 & Global \\
Death Toll               & 2 to 165,000 \\
Economic Damage          & \$0.12 to 360 billion \\
Living Space             & villages to major cities \\
Resolution               & 0.3m to 3m \\
\end{tabular}
\end{table}

\begin{table}
\small
\centering
\caption{Baseline ResNet-34 for Robust Disaster Detection: Architecture and Training/Testing Parameters}
\label{Table2}
\begin{tabular}{ccc}
\hline
Layer       & Output Size       & Details \\ \hline
Conv 1      & 112$\times$112    & 7$\times$7, 64, stride 2 \\
Conv 2      & 56$\times$56      & 3$\times$3 max pool, stride 2 \\
            &                   & $[3\times3,64,3\times3,64]\times3$\\
Conv 3      & 28$\times$28      & $[3\times3,128,3\times3,128]\times4$\\ 
Conv 4      & 14$\times$14      & $[3\times3,128,3\times3,256]\times6$\\ 
Conv 5      & 7$\times$7        & $[3\times3,128,3\times3,512]\times3$\\  
Pool, FC  & 1$\times$1        & Average, 1000-d FC, SoftMax \\ \hline
Training    &                   & Value \\ \hline
Batch Size  &                   & 16 \\
Epochs      &                   & 30 \\
Learning Rate   &               & 0.00001 \\
Loss Func.  &                   & Cross Entropy \\
Optimizer   &                   & Adam \\
FLOPs       &                   & $3.6\times10^{9}$ \\
Test Performance &              & 92\% Accuracy, 0.05 Loss \\
\end{tabular}
\end{table}

\begin{table}
\small
\centering
\caption{PSPNet for Scene Parsing Disaster Scale Quantification through Build Environment Segmentation}
\label{Table3}
\begin{tabular}{cc}
\hline
Training    & Value \\ \hline
Backbone    & DenseNet \cite{DenseNet} \\
Batch Size  & 4 \\
Epochs      & 100 \\
Learning Rate   & 0.0001 \\
Loss Func.  & Cross Entropy \\
Optimizer   & Adam \\
Test Performance & 84-88\% Accuracy, 0.21 Loss \\
\end{tabular}
\end{table}

\begin{figure*}[t]
	\centering
	\includegraphics[width=1.0\linewidth]{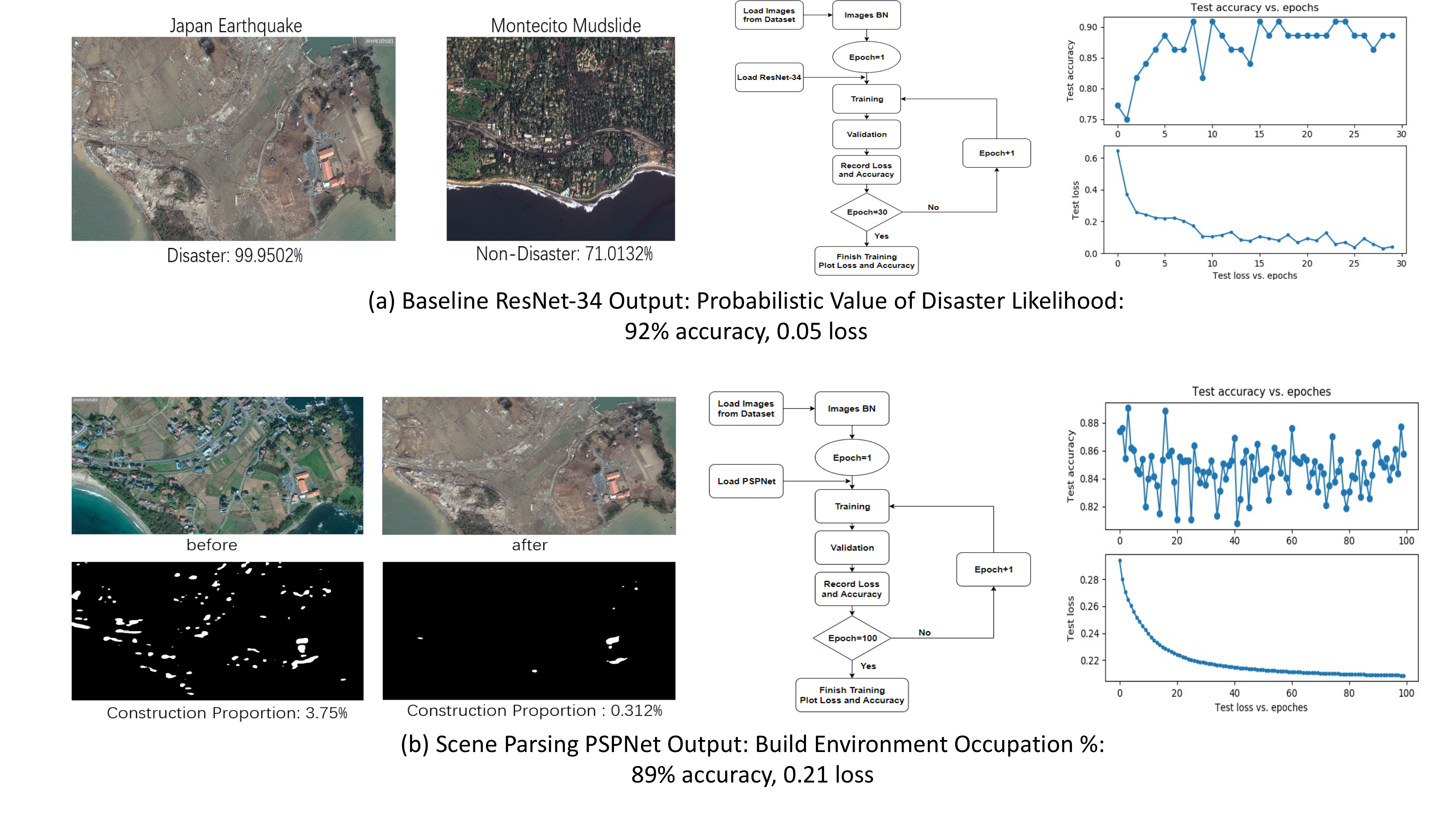}
	\caption{Baseline ResNet and Scene Parsing PSPNet: data, training, and performance. (a) ResNet produces a probability score for a disaster with high accuracy and is robust to noise, whereas (b) PSPNet can quantify the amount of built environment damage with a lower accuracy and is not robust to noise.}
	\label{2}
\end{figure*}

\begin{table*}[t]
\centering
\caption{Case Study Results on Best and Worst Performance}
\label{Table4}
\begin{tabular}{ccccc}
\hline
Location    & Disaster          & Damage            & ResNet    & PSPNet \\ \hline
Japan       & Earthquake (2011) & \$369bn 19k death & 100\%     & 90\% \\
Indonesia   & Tsunami (2004)    & \$94bn 16k death  & 99\%     & 80\% \\
Oklahoma    & Tornado (2013)    & \$20bn 29 death   & 90\%     & 65\% \\ 
Philippines & Typhoon (2013)    & \$30bn 7k death   & 69\%     & 71\% \\  
Guatemala   & Volcano (2018)    & \$0.12bn 461 death& 69\%     & 45\% \\  
Montecito   & Mudslide (2018)   & \$2bn 21 death    & 63\%     & 59\% \\ 
Bahamas     & Hurricane (2019)  & \$4.7bn 370 death & 55\%     & 40\% \\  
Texas       & Hurricane (2017)  & \$125bn 88 death  & 52\%     & 64\% \\ 
Malibu      & Fire (2018)       & \$6bn 2 death     & 51\%     & 46\% \\  
Florida     & Hurricane (2018)  & \$25bn 74 death   & 51\%     & 63\% \\  
\end{tabular}
\end{table*}

\begin{figure*}[t]
	\centering
	\includegraphics[width=0.9\linewidth]{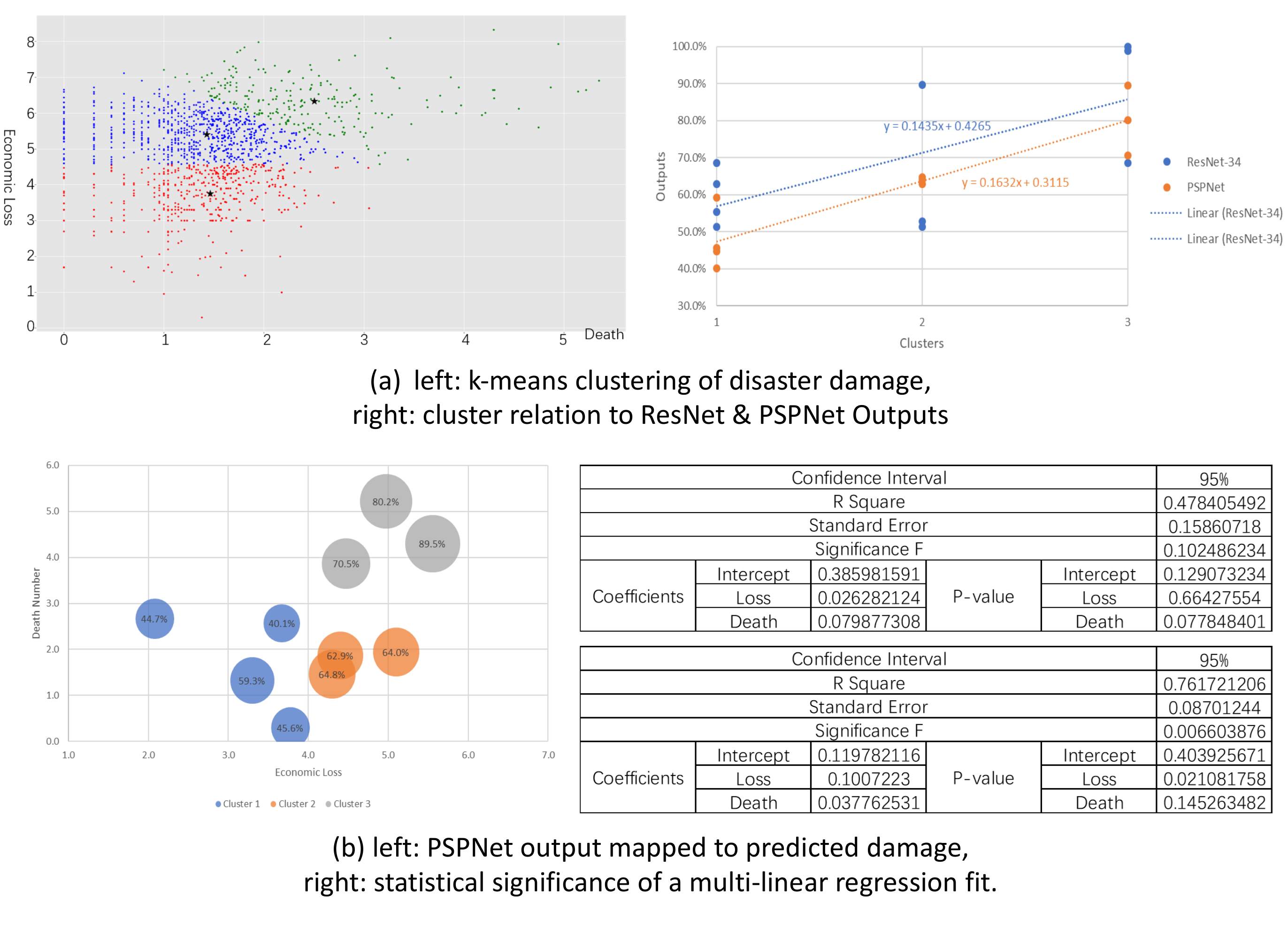}
	\caption{Prediction of Severity and Damages using Unsupervised Clustering and a Multi-linear regression linking CNN outputs with Damage Data.}
	\label{3}
\end{figure*}

\begin{figure}[t]
	\centering
	\includegraphics[width=0.9\linewidth]{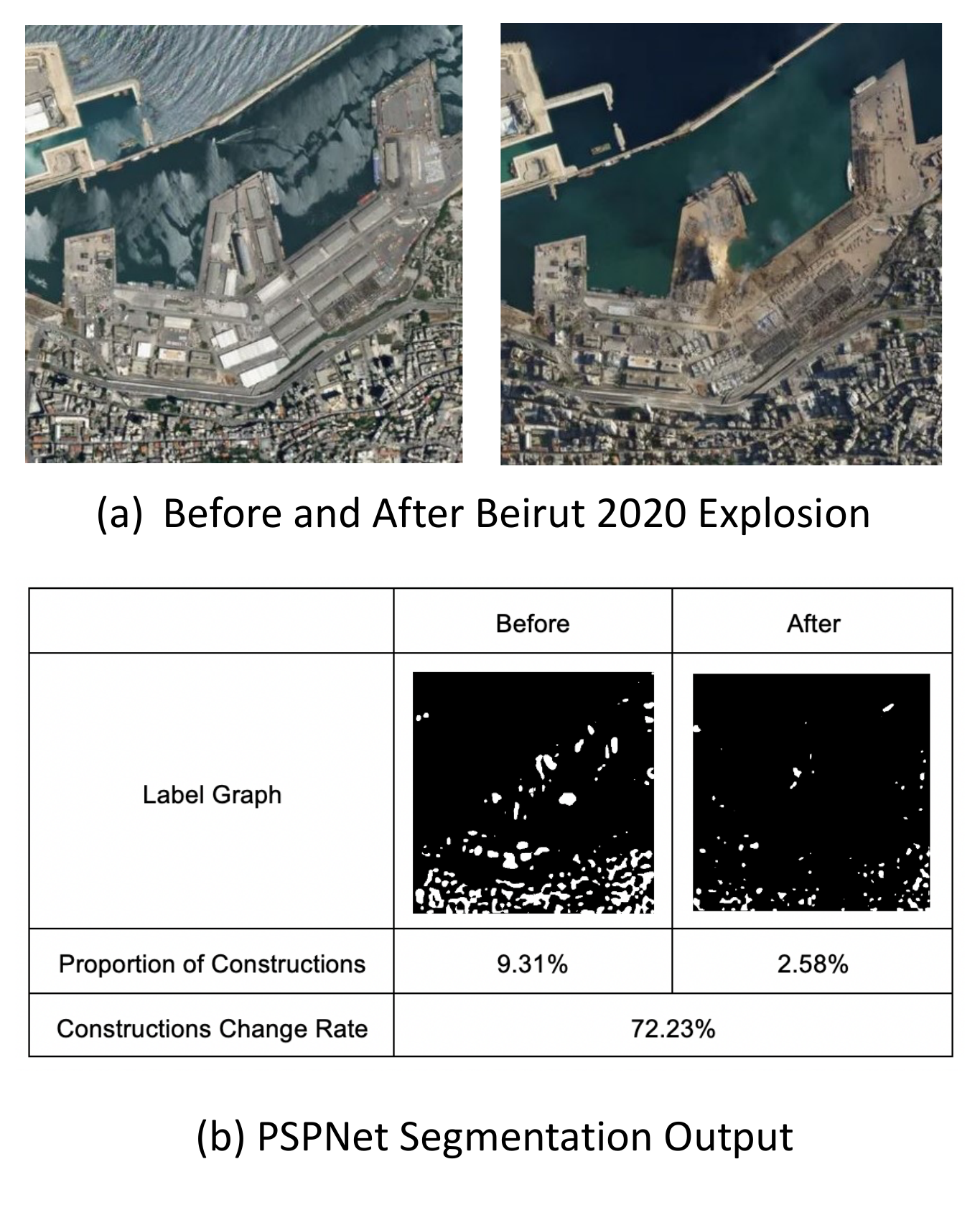}
	\caption{Case Study on Beirut Port Explosion 2020: (a) raw photos, and (b) PSPNet Output and Predicted Damage.}
	\label{4}
\end{figure}

\subsection{ResNet for Baseline Classification}
We first consider the widely used ResNet model \cite{ResNet}, which is designed to get an initial output for the severity of disasters according to the images for pre- and post-disaster. ResNet has the advantage of being relatively robust as poor image quality and different noises. Noises can often arise due to environmental conditions (e.g. smoke from conflict or fire) or compression of image due to communication bandwidth and edge processing limits. As a baseline, we first attempt a binary ResNet model with two classifications, disaster and non-disaster, and the output is the final classification with a probability based on training. A total of 271 images was used for training and 44 images used for validation. In the ResNet-34 model, 34 convolutional layers are used and the architecture used is given in Table 2. The baseline ResNet (see Figure 2a) is robust to noise (loss 0.05) and has a high accuracy of 92\%, but it can only quantify the likelihood of a disaster present. This is sufficient for prioritising further investigation on a large scale, but is not effective in quantifying the magnitude of the damage as it cannot classify the built environment and assign a notion of contextualised value or worth.

\subsection{PSPNet for Scene Parsing}
Whilst the baseline ResNet is robust to noise and has a high accuracy, it can only quantify the likelihood of a disaster present. PSPNet is therefore selected and configured, because it can segment all build environments and label them for quantification later in the paper \cite{PSPNet}. The proportion of constructions area in the satellite images before and after disasters can be calculated. Through these two data, the change rate of constructions can be obtained. From the change rate of constructions, disaster situations can be reflected quantitatively. A total of 661 images was used for training and 59 images used for validation. In the PSPNet model, a backbone is based on DenseNet \cite{DenseNet}. A feature map CNN (see Figure 1) of 3 layers pooled at 4 different sizes. After which they are convolved with $1\times1$ filters to reduce the depth of the feature. Next, all the features are up-sampled and concatenated. The model of PSPNet also has been created and pre-trained by Pytorch and the process of training model is displayed in Figure 1 with architecture in Table 2.

Using the output segmentation (white colour indicates healthy built environment), we are able to identify the volume of change after a disaster. The scene parsing PSPNet (see Figure 2b) is not as robust to noise (loss 0.21) as ResNet, and has a lower accuracy of 84-88\%, but it can quantify the damage to built environment from a disaster. This enables us to attempt to quantify the economic and human cost of at post disaster analysis or during an unfolding disaster. 

\section{Results for Disaster Cost Quantification}

\subsection{Disaster Case Studies}
The outputs of ResNet-34 and PSPNet model are listed in Table 4 for a few selected case study in different areas of the world, under different type and scale disasters, with varying levels of damage. From the table, it can be seen that the difference between ResNet-34 and PSPNet has an average of 8\%. However, we can see that both are quite accurate (80-100\%) in predicting the disaster and level of damage for large disasters (e.g. Japanese 2011 earthquake or Indonesia 2004 tsunami), but has poor accuracy for small disasters (e.g. Malibu fire 2019 or Florida hurricane 2018). This is partly because of the lack of clear differentiation in damage in smaller disasters, as well as the lower volume of training data (e.g. training is biased towards larger data sets).

\subsection{Projecting the Economic and Life Loss}
Previously the CNNs were able to identify which features mapped to damages, but could not yet appropriate a economic and human life cost to the damages. The 2 primary damage labels for every event are economic loss ($\$bn$) and the number of deaths. We wish to cluster them into a number of severity categories in order to reduce the resolution and dimensionality of the problem into a single \textit{severity class} scale. 

We select unsupervised $k$-means to cluster over 2000 cases of natural disasters from 2000 to 2019 using data from EM-DAT \cite{EM-DAT}. From the Figure 3, it can be found that all data has been divided into three clusters, a different colour distinguishes every cluster of severity. This analysis is relatively reasonable because three cluster centres (three stars in Figure 3a-left) show the linear relationship, it is smooth and clear to judge red, blue and green area in the graph representing gentle, medium and severe disasters respectively. The number of cluster 1 cases is 779, which is 39\% of all cases, and proportions of cluster 2 and cluster 3 are 32\% and 29\%. 

In Figure 3a-right, we put every event into the corresponding cluster, and then plot outputs corresponding to every event into a graph as shown in Figure 3-2 with both ResNet-34 and PSPNet outputs, the linear relationships relate CNN outputs with the actual damage scale. The confidence interval of both models is set to 95\% (see Figure 3b). It can be found that the R2 of ResNet-34 model is 0.48, which is much smaller than that of PSPNet at 0.76. Our multi-linear regression model relating PSPNet Loss output to predicted damages is:
\begin{equation}
    \text{Loss} = 0.1 \times \text{Economic Loss} + 0.038 \times \text{Deaths} + 0.12,
\end{equation} and we have checked that the residue is normal distributed.

\subsection{Case Study: Beirut Port Explosion 2020}
In 4th August of 2020, there was a powerful explosion in the port warehouse area of the Lebanese capital Beirut, causing widespread damage to the capital and destroying almost all buildings near the sea. At least 154 people have been killed, and nearly 5000 injured in the explosions (The Guardian, 2020). The main effect of the explosion radiated out from the point of explosion, so it is unnecessary to input wider satellite images of distant Beirut areas and only a single port area satellite image is used. First, we put these two images (Figure 4a) imported into the database with some necessary information such as location, date, and number of deaths. The economic loss is uncertain now, so the purpose is to find an approximate economic loss of this disaster. Second, using the pre-trained PSPNet model, we predict the damage in Figure 4b. The predicted loss in built livable space 72\% under the severe category. Using Equation 1, the predicted economic loss is \$15.6bn US dollars. This corresponds with early government estimates of \$10-15bn. As such, despite large explosions were not part of the training data, the PSPNet model can be successfully used in disaster management to predict and assess disaster costs.

\section{Conclusions}

Humanitarian disasters and political violence cause significant damage to our living space. The reparation cost to built livable space (e.g. homes, infrastructure, and the ecosystem) is often difficult to quantify in real-time. Real-time quantification is critical to both informing relief operations, but also planning ahead for rebuilding. Here, we used satellite images before and after major crisis around the world for the last 20 years to train a new Residual Network (ResNet) and Pyramid Scene Parsing Network (PSPNet) to quantify the magnitude of the damage. ResNet offers the robustness to poor image quality, whereas PSPNet offers scene parsing contextualised analysis of damage. 

Both of these techniques are useful and can be cascaded: (Step 1) ResNet can identify priority areas on low resolution imagery over a large area with 90\% accuracy, and this can be followed up with (Step 2) PSPNet to accurately identify the level of damage with 80\% accuracy. As there are multiple damage dimensions to consider (e.g. economic loss, death-toll), we fitted a multi-linear regression model to quantify the overall damage cost to the economy and human lives. To validate our model, we successfully match our prediction to the ongoing recovery in the 2020 Beirut port explosion. These innovations provide a better quantification of overall disaster magnitude and inform intelligent humanitarian systems of unfolding disasters.

\bibliographystyle{IEEEtran}
\bibliography{IEEEabrv,Ref}

\begin{thebibliography}{10}
\providecommand{\url}[1]{#1}
\csname url@samestyle\endcsname
\providecommand{\newblock}{\relax}
\providecommand{\bibinfo}[2]{#2}
\providecommand{\BIBentrySTDinterwordspacing}{\spaceskip=0pt\relax}
\providecommand{\BIBentryALTinterwordstretchfactor}{4}
\providecommand{\BIBentryALTinterwordspacing}{\spaceskip=\fontdimen2\font plus
\BIBentryALTinterwordstretchfactor\fontdimen3\font minus
  \fontdimen4\font\relax}
\providecommand{\BIBforeignlanguage}[2]{{%
\expandafter\ifx\csname l@#1\endcsname\relax
\typeout{** WARNING: IEEEtran.bst: No hyphenation pattern has been}%
\typeout{** loaded for the language `#1'. Using the pattern for}%
\typeout{** the default language instead.}%
\else
\language=\csname l@#1\endcsname
\fi
#2}}
\providecommand{\BIBdecl}{\relax}
\BIBdecl

\bibitem{4215094}
S.~{Voigt}, T.~{Kemper}, T.~{Riedlinger}, R.~{Kiefl}, K.~{Scholte}, and
  H.~{Mehl}, ``Satellite image analysis for disaster and crisis-management
  support,'' \emph{IEEE Trans. on Geoscience and Remote Sensing}, vol.~45,
  no.~6, 2007.

\bibitem{7892890}
X.~{Chen}, J.~{Li}, Y.~{Zhang}, W.~{Jiang}, L.~{Tao}, and W.~{Shen},
  ``Evidential fusion based technique for detecting landslide barrier lakes
  from cloud-covered remote sensing images,'' \emph{IEEE Journal of Selected
  Topics in Applied Earth Observations and Remote Sensing}, vol.~10, no.~5, pp.
  1742--1757, 2017.

\bibitem{8114334}
I.~{Coulibaly}, N.~{Spiric}, R.~{Lepage}, and M.~{St-Jacques}, ``Semiautomatic
  road extraction from vhr images based on multiscale and spectral angle in
  case of earthquake,'' \emph{IEEE Journal of Selected Topics in Applied Earth
  Observations and Remote Sensing}, vol.~11, no.~1, pp. 238--248, 2018.

\bibitem{8585069}
J.~{Wang}, K.~{Sato}, S.~{Guo}, W.~{Chen}, and J.~{Wu}, ``Big data processing
  with minimal delay and guaranteed data resolution in disaster areas,''
  \emph{IEEE Transactions on Vehicular Technology}, vol.~68, no.~4, pp.
  3833--3842, 2019.

\bibitem{Amit16}
S.~Amit, S.~Shiraishi, T.~Inoshita, and Y.~Aoki, ``Analysis of satellite images
  for disaster detection,'' \emph{IEEE International Geoscience and Remote
  Sensing Symposium}, 2016.

\bibitem{RSOS}
A.~Pamuncak, W.~Guo, A.~Soliman, and I.~Laory, ``Deep learning for bridge load
  capacity estimation in post-disaster and -conflict zones,'' \emph{Royal
  Society Open Science}, vol.~6, no.~12, p. 6190227, 2019.

\bibitem{Kamilaris18}
A.~Kamilaris and F.~Prenafeta-Boldú, ``{Disaster Monitoring using Unmanned
  Aerial Vehicles and Deep Learning},'' \emph{CoRR}, 2018.

\bibitem{Duarte18}
D.~Duarte, F.~Nex, N.~Kerle, and G.~Vosselman, ``Satellite image classification
  of building damages using airborne and satellite image samples in a deep
  learning approach,'' \emph{ISPRS Annals of the Photogrammetry, Remote Sensing
  and Spatial Information Sciences}, 2018.

\bibitem{Sublime19}
J.~Sublime and E.~Kalinicheva, ``Automatic post-disaster damage mapping using
  deep-learning techniques for change detection: Case study of the tohoku
  tsunami,'' \emph{Remote Sensing}, 2019.

\bibitem{Guo19}
W.~Guo, K.~Gleditsch, and A.~Wilson, ``Retool ai to forecast and limit wars,''
  \emph{Nature}, vol. 562, pp. 331--333, 2019.

\bibitem{EM-DAT}
``{EM-DAT (2020) Database},'' 2020.

\bibitem{Maggiori17}
E.~Maggiori, Y.~Tarabalka, G.~Charpiat, and P.~Alliez, ``Can semantic labeling
  methods generalise to any city? the inria aerial image labeling benchmark,''
  \emph{Int. Geoscience and Remote Sensing Symposium}, 2017.

\bibitem{Gupta19}
R.~Gupta, R.~Hosfelt, S.~Sajeev, N.~Patel, B.~Goodman, E.~D. adn E.~Heim,
  H.~Choset, and M.~Gaston, ``xbd: A dataset for assessing building damage from
  satellite imagery,'' \emph{IEEE Conference on Computer Vision and Pattern
  Recognition (CVPR)}, 2019.

\bibitem{DenseNet}
G.~Huang, Z.~Liu, L.~V.~D. Maaten, and K.~Weinberger, ``Densely connected
  convolutional networks,'' \emph{IEEE Conference on Computer Vision and
  Pattern Recognition (CVPR)}, 2017.

\bibitem{ResNet}
K.~He, X.~Zhang, S.~Ren, and J.~Sun, ``Deep residual learning for image
  recognition,'' \emph{IEEE Conference on Computer Vision and Pattern
  Recognition (CVPR)}, 2016.

\bibitem{PSPNet}
H.~Zhao, J.~Shi, X.~Qi, X.~Wang, and J.~Jia, ``Pyramid scene parsing network,''
  \emph{IEEE Conference on Computer Vision and Pattern Recognition (CVPR)},
  2017.

\end{thebibliography}



\end{document}